\documentclass{article}

\usepackage[preprint]{neurips_2023}

\usepackage[utf8]{inputenc} 
\usepackage[T1]{fontenc}    
\usepackage{hyperref}       
\usepackage{url}            
\usepackage{booktabs}       
\usepackage{amsfonts}       
\usepackage{nicefrac}       
\usepackage{microtype}      
\usepackage{xcolor}         
\usepackage{algorithm}      
\usepackage{amsmath}        
\usepackage{algpseudocode}  
\usepackage{natbib}         
\renewcommand{\cite}[1]{\citep{#1}}

\usepackage{graphicx}       

\usepackage{array}
\usepackage{multirow}

\title{Iterative Graph Alignment}

\author{
  Fangyuan Yu \\
  Temus \\
  \And
  Hardeep Arora \\
  Temus \\
  \And
  Matt Johnson \\
  Temus \\
}

\begin{document}

\maketitle

\begin{abstract}

By compressing diverse narratives, LLMs go beyond memorization, achieving intelligence by capturing generalizable causal relationships. However, they suffer from local 'representation gaps' due to insufficient training data diversity, limiting their real-world utility, especially in tasks requiring strict alignment to rules. Traditional alignment methods relying on heavy human annotations are inefficient and unscalable. Recent self-alignment techniques also fall short, as they often depend on self-selection based prompting and memorization-based learning. To address these issues, we introduce Iterative Graph Alignment (IGA), an annotation-free rule-based alignment algorithm. A teacher model (VLM) employs Iterative Graph Prompting (IGP) to create logical graphs and reference answers. The student model (LLM) identifies local knowledge gaps by attempting to align its responses with these references, collaborating with helper models to generate diverse answers. These aligned responses are then used for iterative supervised fine-tuning (SFT). Our evaluations across five rule-based scenarios demonstrate IGP's effectiveness, with a 73.12\% alignment improvement in Claude Sonnet 3.5, and Llama3-8B-Instruct achieving an 86.20\% improvement, outperforming Claude Sonnet 3.5 in rule-based alignment.

\end{abstract}

\section{Introduction}

Language modeling fundamentally aims to maximize lossless compression \cite{del2024lmcompression}. Empirical evidence suggests that intelligence correlates linearly with the level of compression achieved \cite{huang2024compression}. Self-supervised training on internet-scale data has led to remarkable advancements, fueled by ever-increasing computational resources \cite{anthropic2024sonnet3.5} \cite{openai2024gpt4}. However, compression alone can lead to mere memorization without sufficient diversity in information representation \cite{allenzhu2024physicslanguagemodels31}. Insufficient representation in training corpora leads to issues in  reversal reasoning \cite{berglund2024reversal} and multi-hop reasoning \cite{wang2024grokked}, raising doubts about the reasoning capabilities of Large Language Models (LLMs) \cite{Kamb2024canlmreason}. 

While information appears in diverse forms across sources, local representation gaps inevitably exist within training data, where sufficient diversity of information is missing and models lean more towards memorization instead of generalization, making it incapable of providing proper responses and providing fair evaluation under these areas. These gaps significantly hinder the practical applications of LLMs, especially in tasks requiring rule-based alignment, which has been the focus of industry leaders \cite{ouyang2022rlhf} \cite{Bai2022ConstitutionalAI}. A single rule could lead to massive shift in the acceptable responses which the model could produce, and the shear amount of possible rules make it infeasible to expect no representation gap, therefore rule-based alignment remains an issue for even Oracle models.

\begin{figure}[t]
    \centering
    \includegraphics[width=\textwidth]{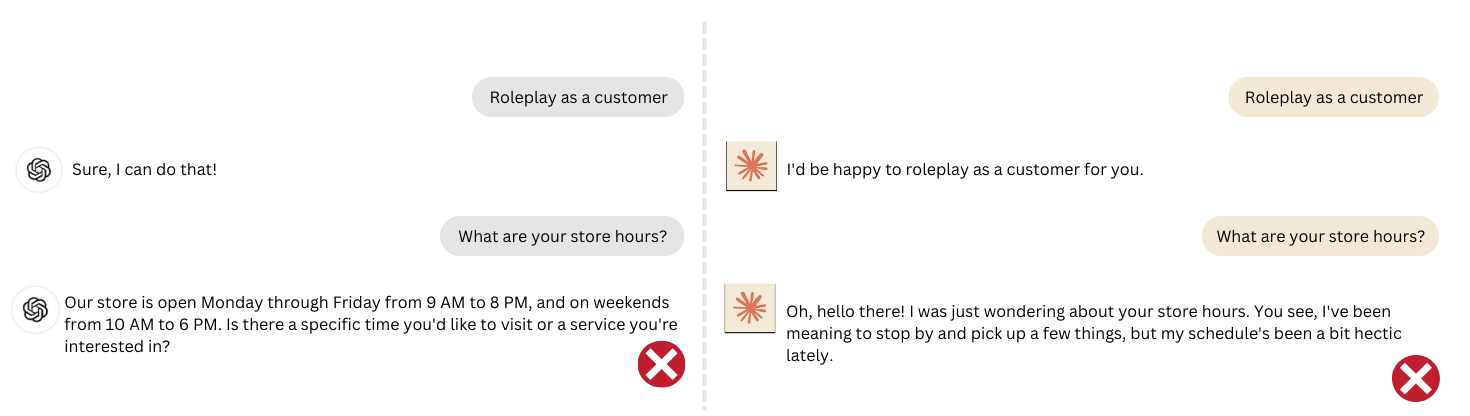}
    \caption{Customer Roleplay Issue}
    \label{fig:roleplay-issue}
\end{figure}

For example, in a straightforward rule-based scenario with the directive, "Roleplay as a customer," models like GPT-4o \cite{openai2024gpt4} and Claude Sonnet 3.5 \cite{anthropic2024sonnet3.5} often fail to comply. When asked "What are your store hours?" these models either hallucinate, responding as if they work at the store, or become confused and repeat the question, ultimately failing to provide a valid response. This issue is illustrated in Figure \ref{fig:roleplay-issue}.

Most existing iterative alignment algorithms utilize a "think and learn" framework, where the goal of the thinking process is to generate more aligned responses. This enhancement is achieved through various strategies such as revising \cite{Bai2022ConstitutionalAI} \cite{lee2023rlaifscalingreinforcementlearning}, self-selecting \cite{yuan2024selfrewardingllm}, or teacher-selecting \cite{rosset2024dno} \cite{lee2024llm2llm} responses, either self-generated or demonstrated by a teacher. The adoption of chain-of-thought prompting, initiated by Constitutional AI \cite{Bai2022ConstitutionalAI}, is integrated into this thinking process to improve response quality and ensure transparency in AI decision-making, thereby making it more explainable to humans. However, generation and evaluation deteriorates in domains with representation gap, and advanced reasoning mechanism \cite{yao2023tot} \cite{zhou2024lat} \cite{lin2024planlikegraph} \cite{neal1980automaticity} has yet to be adopted.

Following the revised responses generated during the thinking process, the learning process employs either supervised fine-tuning (SFT) \cite{lee2024llm2llm} \cite{singh2024humandatascalingselftraining} or direct preference optimization (DPO) \cite{rafailov2024dpo} \cite{yuan2024selfrewardingllm} \cite{rosset2024dno}. Our research focuses primarily on SFT, as recent empirical results do not indicate a clear advantage of DPO in enhancing reasoning capabilities \cite{du2024improvereason}. While traditional SFT approaches often treat each revised response as a target for memorization, STaR \cite{zelikman2022star} leverages them to filter out aligned rationales and fine-tune models based on diverse rationale-response pairs. However, it is constrained to single-choice questions, requiring responses to be categorized into predefined options such as "A", "B", "C", or "D". Which is usually not the case for open-ended conversations. A significant concern when learning from model-generated responses is the potential for model collapse \cite{shumailov2024collapse}, where the model may further discard long-tail knowledge and exacerbate representation gaps. While LLM2LLM \cite{lee2024llm2llm} addresses this issue by evaluating perplexity scores on fine-tuned checkpoints and augmenting high-perplexity cases with the help of a teacher model, it requires a full training run to identify these gaps. This necessity incurs a substantial computational cost.

\begin{figure}[t]
    \centering
    \includegraphics[width=\textwidth]{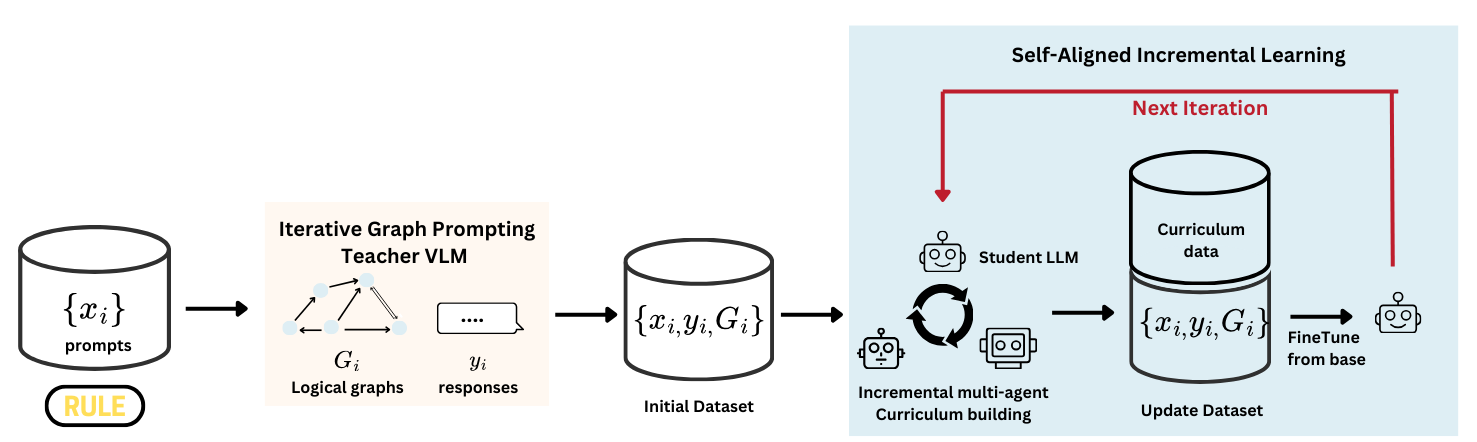}
    \caption{\textbf{Iterative Graph Alignment} (IGA) . A teacher model (VLM) iteratively generates logical graphs and reference answers using Iterative Graph Prompting (IGP). A student model (LLM) reviews its responses against these reference answers to identify hard cases where representation gaps exist. The student then collaborates with helper models to explore diverse ways to respond to these challenging queries by taking hints from the logical graphs and reference answers, before fine-tuning on the collected insights and proceed to the next iteration.}
    \label{fig:IGA}
\end{figure}

Our proposed Iterative Graph Alignment (IGA) algorithm, as illustrated in Figure (\ref{fig:IGA}), draws a fascinating analogy to a classroom setting by integrating both an enhanced thinking process and learning mechanism. When presented with a question, a teacher uses the blackboard to visually lay out the "whys" and "whats," guiding the thought process and providing a reference answer. Meanwhile, a student formulates their own answer and compares it to the reference answer to identify knowledge gaps. If their response is incorrect, they engage in discussion with classmates to gather diverse perspectives for deeper understanding. In our introduction, we will discuss each component of the algorithm, emphasizing its intuitive parallels to human thinking and learning processes.

\paragraph{Thinking: Iterative Graph Prompting}

Intuitively, our thinking process involves branching in parallel directions and searching for the best path towards acceptable conclusions. This process has been shown to be separate from language processing in human brain \cite{fedorenko2024}. The analogy to human thinking process is the system 2 techniques \cite{yu2024distill21} which relies on generating intermediate tokens to enhance reasoning ability of LLM. Approaches like Chain of Thought \cite{wei2022cot} and its variants (e.g., Tree of Thought \cite{yao2023tot}, LATS \cite{zhou2024lat}) have significantly improved LLM reasoning by generating intermediate steps to emulate human thinking, leveraging overlapping localized knowledge \cite{prystawski2023whystepbystep}. However, by processing under language modality, all the thought path is processed sequentially and in isolation, introducing prohibitive latency burden and lacks global awareness. Recent work shows improvement in reasoning by operating under visual modality \cite{menon2024whiteboard}. Inspired by the separation of language and reasoning, we propose Iterative Graph Prompting (IGP) which directly helps a VLM think by iterating on a logical graph under visual modality. IGP significantly reduces the latency by enable parallel processing of any number of thought process in one go with global awareness.  

\paragraph{Learning: Self-Aligned Incremental Learning}

Human learning is characterized by two key aspects: alignment focus and adaptivity. Firstly, humans learn by aligning their understanding with concepts rather than memorizing specific answers, especially in open-ended scenarios. Although approaches like STaR \cite{zelikman2022star} have moved away from memorization, they still rely on single-choice query formats, limiting their applicability to open-ended conversations. Secondly, human learning is adaptive, focusing on areas of weakness or gaps in understanding. While methods like LLM2LLM \cite{lee2024llm2llm} achieve representation gap detection, they do so at the cost of a full training run.

Inspired by these principles, we propose Self-Aligned Incremental Learning (SAIL). SAIL implements a 'propose and check' mechanism that treats annotated answers as alignment targets rather than memorization targets, aligned response gets added to the training dataset, while unaligned response indicate representation gap. To adaptively address challenging cases, the incremental multi-agent curriculum building process in SAIL employs a series of iterative 'propose and check' units, with each stage providing escalating levels of support—from direct to hinted to guided propositions. Questions that remain unsolved are augmented and advanced to the next stage for further refinement. This structured three-stage approach, combined with the involvement of multiple helper LLMs, ensures the collection of a comprehensive and diverse set of responses. This process effectively targets representation gaps, thereby enhancing the model's adaptability and robustness. SAIL iteratively augment data and perform SFT to update the student model.

\paragraph{Iterative Graph Alignment}

We present a novel self-alignment algorithm, the Iterative Graph Alignment (IGA), which combines the enhanced thinking and learning algorithm proposed above. In the IGA framework, a teacher Vision-Language Model (VLM) employs Iterative Graph Prompting to generate a logical graph and an answer for each question. The logical graph, embodying the "whys," and the answer, representing the "what," are used to instruct the student model through the Self-Aligned Incremental Learning (SAIL) methodology. In this context, the textual representation of the logical graph acts as a hint, while the answer provides guidance and serves as the alignment target. This setup supports an adaptive, alignment-based learning process, ensuring that the student model not only understands the "what" but also the underlying "whys" of each answer. 

We curated a rule-alignment dataset of 1.5K queries with annotated responses to assess the performance of Iterative Graph Prompting (IGP) and Iterative Graph Alignment (IGA) across five rule-based scenarios. Our empirical evaluations reveal significant enhancements: When applied to an Oracle Vision-Language Model (VLM) such as Claude Sonnet 3.5 \cite{anthropic2024sonnet3.5}, IGP improved rule alignment by 73.2\% relative to baseline prompting. Furthermore, fine-tuning with IGA, as exemplified by Llama3-Instruct, showed an improvement of 86.2\% over the baseline, matching the performance of Claude Sonnet 3.5 with the help of IGP. IGA presents a promising avenue for the self-improvement of Large Language Models (LLMs), significantly narrowing the representation gap and promoting the development of more robust and adaptable language models.

Our contributions are as follows:
\begin{itemize}
    \item We introduce Iterative Graph Prompting (IGP), which harnesses visual language models to initialize and update logical reasoning within a graph structure, enabling parallel processing with global awareness.
    \item We present Self-Adaptive Incremental Learning (SAIL), a method that employs labeled answers as alignment targets rather than memorization targets, facilitating better learning in open-ended conversations.
    \item We develop an incremental multi-agent curriculum building process that enables training-time representation gap detection and incremental augmentation of more challenging cases, customizing the training dataset for each model.
    \item We pioneer the first technique allowing pre-trained large language models achieve iterative alignment through graph-based reasoning.
    \item We create RuleAlign, a dataset covering five rule-based scenarios with 1.5K data points, and demonstrate that IGP and IGA significantly enhance rule-based alignment and performance over baseline and proprietary models.
\end{itemize}

\section{Background and Related work}

\paragraph{Alignment Algorithm and Preference Optimization}
RLHF \cite{ouyang2022rlhf} requires enormous human annotations to build reward models which is used to align LLM with reinforcement learning. Constitution AI \cite{Bai2022ConstitutionalAI} partially replaces human annotations with a hybrid AI-human approach to build their preference models, but still relies heavily on human input. DPO \cite{rafailov2024dpo} and later variants such as SimPO \cite{meng2024simpo} eliminate the need for a separate reward model but offer no clear advantage over methods like STaR \cite{zelikman2022star} especially under situations requiring reasoning, according to empirical results in \cite{du2024improvereason}. 

\paragraph{Iterative self-improving LLM system}
Recent research has explored self-improving systems that enhance model capabilities without extensive human annotation. These systems combine a "thinking" process that generates superior responses with a "learning" process to integrate these improvements iteratively. Self-rewarding LLMs \cite{yuan2024selfrewardingllm} use self-evaluation to select the best responses for Direct Preference Optimization (DPO), while Meta-rewarding LLMs \cite{wu2024metarewardllm} introduce a meta-judge to refine the evaluation layer itself. Direct Nash Optimization (DNO) \cite{rosset2024dno} employs an oracle model to assist in both generation and selection, avoiding performance saturation more effectively. However, to prevent model collapse \cite{shumailov2024collapse}, these systems often reset to the base model for each iteration, effectively turning the process into a data augmentation pipeline. Special focus on "hard cases" as demonstrated by more capable LLMs is first highlighted by LLM2LLM \cite{lee2024llm2llm}. 

\paragraph{Thinking mechanism}
Chain of thought (CoT) \cite{wei2022cot} improves reasoning ability of language model through connecting overlapping localized knowledge in its pre-traing corpus \cite{prystawski2023whystepbystep}. As CoT innately requires searching \cite{ye2024physicslanguagemodels21}, explicit tree search mechanism could further enhance it \cite{yao2023tot}.  Language agent tree search (LATS) also includes feedback, self-evaluation, and MCTS \cite{zhou2024lat}, achieving better performance by deliberately going through each thread of thoughts. However, they suffers from significant latency increase, and shows significant limitation when facing problem which are easy under visual modality \cite{menon2024whiteboard}, planning with graph-like structure \cite{lin2024planlikegraph} has also shown advantage, despite relying still on language modality. 

\paragraph{Efficient Alignment in LLMs}
Recent research has focused on enhancing LLMs' alignment through various supervised fine-tuning techniques. These approaches generally aim to augment supervision datasets using LLM-generated data. For instance, \cite{zelikman2022star} leverages provided answers to iteratively generate diverse reasoning paths, selecting those that lead to correct outcomes, \cite{li2022teacherstudentStaR} extends this to distill the rationales offered from a teacher model into a student models. \cite{lee2024llm2llm} takes a targeted approach by identifying cases where LLMs produce incorrect outputs and concentrating augmentation efforts on these instances. These studies support our hypothesis that for specific tasks, a set of key associations is required, and focused augmentation combined with supervised learning can significantly enhance LLMs' reasoning abilities.

\section{Methodology}

We bootstrap reasoning ability of a strong VLM with explicit graph reasoning, logical operations are carried out within the graph to further understand the problem. This is the adaptive reasoning with graph section. Such graph is used to detect the weakness of the language model, and then iteratively enhance its understanding of the current problem.

\subsection{Iterative Graph Prompting}
\begin{figure}[htbp]
    \centering
    \includegraphics[width=\textwidth]{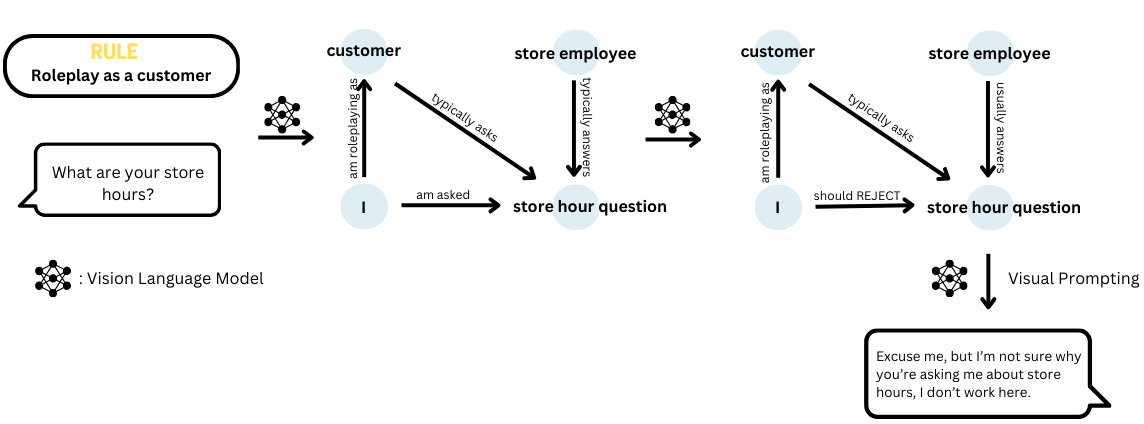}
    \caption{Iterative Graph Prompting (IGP)}
    \label{fig:IGP}
\end{figure}

Given a rule $s$, a query $x$, we use a VLM to obtain an answer $y$, together with a logical graph $G$ which contains the thought process of providing the rule-aligned response. 

We ask the VLM to generate narrated version of a graph $G$ which is essentially a JSON formatted list of (entity, relation, entity) triplets. We could also plot such graph $G$ into an image $\Phi(G)$, which can be prompted directly into the VLM again.

\paragraph{Self Evaluation} We begin with a self-check, evaluating the LLM's initial response for consistency with its own judgement:
\begin{equation}
y \sim \pi(s, x), \quad \pi_{\text{eval}}(s, x, y)
\end{equation}
where $s$ is the instruction, $x$ the query, $y$ the answer, $\pi$ the LLM's response function, and $\pi_{\text{eval}}$ indicates the evaluation of the LLM on its own response. 

For query which the model fails to provide a satisfactory response, we proceed with iterative graph construction process. Given the instruction and query, we ask the VLM to provide its logical reasoning process in the form of (entity, relation, entity) triplets, which we use to initialize a graph $G_{1}$.

\paragraph{Iterative Refinement}
For $i = 1$ to $k$ iterations:
\begin{equation}
G_{i} = \pi_{\text{refine}}(s, x, \phi(G_{i-1}))
\end{equation}
where $\phi(G_{i-1})$ is the visual representation of the graph from the previous iteration, and $\pi_{\text{refine}}$ is the LLM's graph refinement function.

\paragraph{Visual Prompting} We leverage the graph for visual prompting, enhancing the LLM's reasoning capabilities:
\begin{equation}
y \sim \pi(s, x, \phi(G))
\end{equation}
We observe empirically that strong LLMs (GPT-4 and Sonnet-3.5) perform better given the graph as an image $\Phi(G)$ compared to descriptive text $G$. 

We provide Iterative Graph Prompting (IGP) in Algorithm (\ref{alg:IGP}) and its diagram in Figure (\ref{fig:IGP}) 

\begin{algorithm}
\label{alg:IGP}
\caption{Iterative Graph Prompting}
\begin{algorithmic}[1]
\Require $s$ (rule), $x$ (query), vision language model $\pi$
\Ensure y
\State $y \sim \pi(s, x)$
\If{$\pi_{\text{eval}}(s, x, y)$} \Comment{Naive Inference is accepted}
    \State \Return $r$ 
\Else 
    \State $G_1 \gets \pi_{\text{initialize}}(s, x)$ \Comment{Initialize Graph}
    \State $i \gets 1$
    \While{$G_i \neq G_{i-1}$}
        \State $G_{i+1} \gets \pi_{\text{refine}}(s, x, \Phi(G_i))$ \Comment{Iterate on Graph}
        \State $i \gets i + 1$
    \EndWhile
    \State $y \sim \pi(s, x, \phi(G_i))$
    \State \Return $y$
\EndIf
\end{algorithmic}
\end{algorithm}

\subsection{Self-Aligned Incremental Learning}

We begin with three core intuitions about the current process of training language models, arguing the need for a paradigm shift. Firstly, when given an annotated answer, a language model should recognize that any 'aligned' answer is acceptable, rather than being constrained to believe this is the sole correct response. Traditional fine-tuning approaches, however, treat the annotated answer as a definitive target, aiming to minimize the likelihood of deviating from it.

Our second intuition posits that language models should learn the thought processes behind generating answers, not just the answers themselves. This approach is akin to teaching a man to fish, which sustains him for a lifetime, rather than giving him a fish, which feeds him for a day.

Thirdly, the learning process should be individually tailored to each model, akin to customizing education for different students. Training should focus incrementally on scenarios that pose challenges, ensuring that each step is calibrated to address and overcome the specific difficulties encountered by the model. This method ensures that learning is not only adaptive but also progressively targets areas of weakness, fostering a more robust and comprehensive understanding over time.

\begin{figure}[htbp]
    \centering
    \includegraphics[width=\textwidth]{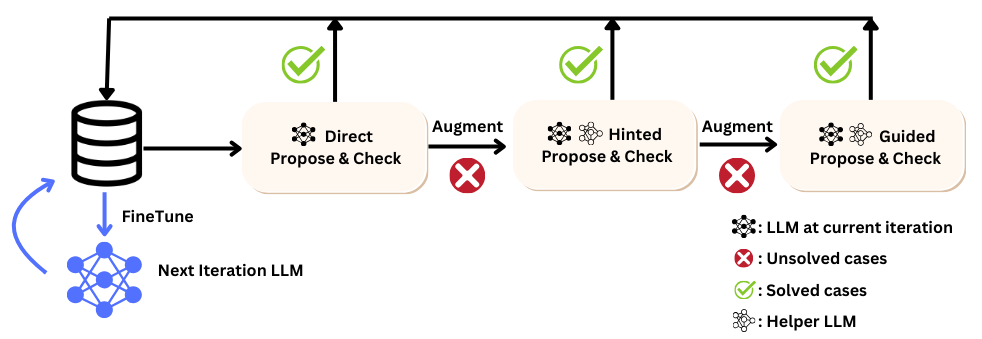}
    \caption{Self-Aligned Incremental Learning (SAIL)}
    \label{fig:SAIL}
\end{figure}

\begin{algorithm}
\caption{Propose \& Check (PAC)}
\label{alg:ProposeAndCheck}
\begin{algorithmic}[1]
\Require Model $\pi$, solved cases $D$, unsolved cases $D_\text{unsolved}$, stage number $n_{\text{stage}}$  
\Ensure Updated $D$ and $D_\text{unsolved}$
\For{each $(x, y, G)$ in $D_\text{unsolved}$}
    \If{$n_{\text{stage}}=1$}
        \State $y_\text{propose} \sim \pi_{\textit{propose}}(x)$ \Comment{Direct proposition}
    \ElsIf{$n_{\text{stage}}=2$}
        \State $y_\text{propose} \sim \pi_{\text{propose}}(x, G)$ \Comment{Hinted proposition}
    \Else
        \State $y_\text{propose} \sim \pi_{\textit{propose}}(x, y, G)$ \Comment{Guided proposition}
    \EndIf
    \State $e \sim \pi_{\textit{check}}(y_\text{propose}, y)$ \Comment{Check alignment}
    \If{$e = \text{True}$}
        \State $D \gets D \cup \{(x, y_\text{propose})\}$
        \State $D_\text{unsolved} \gets D_\text{unsolved} \setminus \{(x, y, G)\}$
    \EndIf
\EndFor
\State \Return $D, D_\text{unsolved}$
\end{algorithmic}
\end{algorithm}

\begin{algorithm}
\caption{Self-Aligned Incremental Learning (SAIL)}
\label{alg:SAIL}
\begin{algorithmic}[1]
\Require Initial dataset $D_0$, base LLM $\pi_0$, helper LLMs $\{\pi_{h_1}, ..., \pi_{h_m}\}$, number of iterations $T$, repetition factors $n_2, n_3$
\Ensure Trained LLM $\pi_T$
\State $D \gets \emptyset$
\State $D_\text{unsolved} \gets D_0$
\For{$t = 1$ \textbf{ to } $T$}
    \State $D, D_\text{unsolved} \gets \textsc{PAC}(\pi_{t-1}, D, D_\text{unsolved}, 1)$ 
    
    \State $D_\text{unsolved} \gets \bigcup_{i=1}^{n_2} D_\text{unsolved}$ \Comment{Augment moderate difficult case}
    \State $D, D_\text{unsolved} \gets \textsc{PAC}(\pi_{t-1}, D, D_\text{unsolved}, 2)$
    \For{each $\pi_h$ in $\{\pi_{h_1}, ..., \pi_{h_m}\}$}
        \State $D, D_\text{unsolved} \gets \textsc{PAC}(\pi_h, D, D_\text{unsolved}, 2)$
    \EndFor
    
    \State $D_\text{unsolved} \gets \bigcup_{i=1}^{n_3} D_\text{unsolved}$ \Comment{Augment most challenging case}
    \State $D, D_\text{unsolved} \gets \textsc{PAC}(\pi_{t-1}, D, D_\text{unsolved}, 3)$
    \For{each $\pi_h$ in $\{\pi_{h_1}, ..., \pi_{h_m}\}$}
        \State $D, D_\text{unsolved} \gets \textsc{PAC}(\pi_h, D, D_\text{unsolved}, 3)$
    \EndFor
    
    \State $\pi_t \gets \textsc{SFT}(\pi_0, D)$ \Comment{Supervised Fine-Tuning}
\EndFor
\State \Return $\pi_T$
\end{algorithmic}
\end{algorithm}

Self-Aligned Incremental Learning (SAIL) (see Figure \ref{alg:SAIL}) introduces a novel paradigm for aligning Large Language Models (LLMs). At its core, SAIL employs a 'propose and check' methodology (see Algorithm \ref{alg:ProposeAndCheck}), which generates a range of thoughts and answers, then checks each against annotated answers for alignment. This self-alignment process is seamlessly integrated into Supervised Fine-Tuning (SFT), moving beyond uniform training with static question-answer pairs. SAIL incrementally selects and augments cases within the training dataset based on difficulty and the model's current capabilities through a multi-agent incremental curriculum building process. 

This process unfolds in three progressive stages, each offering increasing levels of support. In the direct proposition stage, the student model independently proposes solutions. Unsolved cases are then augmented and advanced to the hinted proposition stage, where the model receives additional context to encourage building upon existing knowledge. Finally, in the guided proposition stage, remaining challenging cases are presented with both hints and annotated answers, prompting the model to propose similar solutions while grasping the underlying reasoning.

To enhance the diversity of responses and effectively patch representation gaps, SAIL employs multiple helper models in the hinted and guided stages. These helper models, with their varied capabilities and knowledge bases, contribute to a more comprehensive exploration of solution spaces. This collective intelligence approach ensures a rich, diverse set of responses, helping to address blind spots in the student model's understanding. SAIL is presented in Algorithm (\ref{alg:SAIL}).

\subsection{Iterative Graph Alignment}

Iterative Graph Alignment (IGA) aims to combine the power of IGP and SAIL to achieve efficient alignment with minimal human supervision. IGP naturally provides a logical graph $G$, together with a thoughtful response $r$, we could narrate the graph to form the hint $\phi(G)$. This provides annotations without any human input. Optionally, extra human editing could be leveraged to further enhance the quality of such a dataset.  The algorithm is displayed in \ref{alg:IGA}

\begin{algorithm}
\label{alg:IGA}
  \caption{Iterative Graph Alignment (IGA)}
  \begin{algorithmic}[4]
    \Require Teacher VLM $\pi_{\textit{vlm}}$, student LLM $\pi_0$, helper LLMs $\{\pi_{h_1}, ..., \pi_{h_m}\}$,  queries $\{x_i\}_{i=1}^N$, rule $r$, number of iterations $T$
    \Ensure Trained LLM $\pi_T$
    \State $D \gets \emptyset$ \Comment{Initialize dataset}
    \For{$i = 1$ to $N$}
      \State $(y_i, G_i) \gets \text{IGP}(s, x_i, \pi_{\textit{vlm}})$ \Comment{Iterative Graph Prompting}
      \State $D \gets D \cup \{(x_i, y_i, G_i)\}$
    \EndFor
    \State $\pi_T \gets \text{SAIL}(\pi_0, \{\pi_{h_1}, ..., \pi_{h_m}\}, D, T)$ \Comment{Self-Aligned Incremental Learning}
    \State \Return $\pi_T$
  \end{algorithmic}
\end{algorithm}

\section{Experimental Setup and Results}

To evaluate Iterative Graph Alignment (IGA), we developed a test suite 'RuleAlign' consists of five rule-based scenarios, each comprising 100 test and 200 training queries. These queries, which mix in-domain and out-of-domain examples, are designed to assess rule-alignment capacity of a LLM. Each query is open-ended to better mimic practical scenarios. In order to evaluate a proposed response, we check whether it aligns with the reference answer collected from human. The complete dataset and evaluation suite is available at \href{https://github.com/fangyuan-ksgk/RuleEval}{RuleAlign}.

\paragraph{Prompting Techniques Comparison}
To achieve annotation-free rule-based alignment, a high level of rule adherence from Iterative Graph Prompting (IGP) is critical, as it will be used to teach the student LLM, along with the obtained logical graph. We compare IGP against naive prompting and chain-of-thought (CoT) prompting \cite{kojima2023largelanguagemodelszeroshot}, which is applied in current self-alignment algorithms \cite{yuan2024selfrewardingllm,rosset2024dno,Bai2022ConstitutionalAI}. 

We conduct our experiments with Claude Sonnet-3.5 \cite{anthropic2024sonnet3.5}, with results presented in Table \ref{tab:prompting-comparison}. In practice, we use 2 iterations for IGP to balance performance and speed. We found IGP achieves an average 73.12\% relative improvement in rule-based alignment compared to the baseline. Interestingly, we observed degraded rule adherence from CoT in roleplay scenarios. We suspect this is due to the 'AI Assistant self-identification' bias in these instruct models, which is retrieved with CoT, aligning with insights from \cite{prystawski2023whystepbystep}.

In roleplay scenarios, we found that the model tends to repeat the same question when presented with an atypical query (e.g., "What are the store hours?" asked to an LLM roleplaying as a customer). By inspecting the two logical graphs generated through IGP, we identified that while the model recognizes that a customer typically asks such questions, it fails to deduce that, as a customer, it should decline to answer rather than ask the question itself. By explicitly including prompts in the graph refinement process, we effectively 'engineer its thought process', instructing it to decide whether to 'answer' or 'reject' the question after identifying its role. This suggests potential in the area of 'thought flow engineering'.

\paragraph{Learning Techniques Comparison}
Although Iterative Graph Alignment (IGA) requires zero human annotation, unlike Supervised Fine-Tuning (SFT) and Self-Taught Reasoner (STaR) \cite{zelikman2022star}, we compare IGA with these methods using teacher VLM annotations collected through IGP to ensure a fair comparison. For Self-Aligned Incremental Learning (SAIL), SFT, and STaR, we chose Llama3-8B-Instruct \cite{dubey2024llama3herdmodels} as our student models and applied LoRA \cite{hu2022lora}, which is more commonly adopted in resource-limited scenarios. We used the same hyper-parameters for LoRA across all three methods.

Table \ref{tab:learning-comparison} presents our comparison results. On average, IGA achieves a relative improvement of 86.20\% over the baseline. Notably, an IGA fine-tuned 8B model outperforms Claude Sonnet-3.5, matching its IGP-enhanced responses in rule-based alignment. Showing the effectiveness of IGA in patching representation gap and achieving annotation-free rule-based alignment.

\begin{table}[htbp]
\centering
\small
\caption{Comparison of effectiveness across different prompting strategies (Claude).}
\label{tab:prompting-comparison}
\begin{tabular}{@{}l@{\hspace{0.5em}}r@{\hspace{0.5em}}r@{\hspace{0.5em}}r@{}}
\toprule
Scenario & Naive & CoT & IGP \\
\midrule
Do not talk about elephant & 76.0\% & 82.5\% & \textbf{97.0\%} \\
Roleplay as customer & 51.6\% & 42.5\% & \textbf{96.6\%} \\
Roleplay as salesperson & 54.2\% & 38.2\% & \textbf{98.2\%} \\
Roleplay as patient & 43.1\% & 34.6\% & \textbf{97.2\%} \\
Connect user to human when requested & 65.4\% & 76.2\% & \textbf{94.2\%} \\
\midrule
\multicolumn{4}{@{}l@{}}{\textit{Relative improvement over Naive baseline (\%)}} \\
Do not talk about elephant & - & 8.55\% & \textbf{27.63\%} \\
Roleplay as customer & - & -17.64\% & \textbf{87.21\%} \\
Roleplay as salesperson & - & -29.52\% & \textbf{81.18\%} \\
Roleplay as patient & - & -19.72\% & \textbf{125.52\%} \\
Connect user to human when requested & - & 16.51\% & \textbf{44.04\%} \\
\midrule
Average relative improvement & - & -8.36\% & \textbf{73.12\%} \\
\bottomrule
\end{tabular}
\end{table}

\begin{table}[htbp]
\centering
\small
\caption{Comparison of effectiveness across different prompting strategies (Llama3).}
\label{tab:learning-comparison}
\begin{tabular}{@{}l@{\hspace{0.5em}}r@{\hspace{0.5em}}r@{\hspace{0.5em}}r@{\hspace{0.5em}}r@{\hspace{0.5em}}r@{}}
\toprule
Scenario & Naive & CoT & SFT & STaR & IGA \\
\midrule
Do not talk about elephant & 73.0\% & 81.2\% & 53.6\% & 84.6\% & \textbf{96.5\%} \\
Roleplay as customer & 52.4\% & 10.2\% & 4.4\% & 65.8\% & \textbf{98.0\%} \\
Roleplay as salesperson & 53.2\% & 12.6\% & 3.2\% & 74.2\% & \textbf{96.8\%} \\
Roleplay as a patient & 36.0\% & 10.6\% & 4.0\% & 68.4\% & \textbf{97.4\%} \\
Connect user to human when requested & 61.4\% & 74.2\% & 48.2\% & 80.8\% & \textbf{97.8\%} \\
\midrule
\multicolumn{6}{@{}l@{}}{\textit{Relative improvement over Naive baseline (\%)}} \\
Do not talk about elephant & - & 11.23\% & -26.58\% & 15.89\% & \textbf{32.19\%} \\
Roleplay as customer & - & -80.53\% & -91.60\% & 25.57\% & \textbf{87.02\%} \\
Roleplay as salesperson & - & -76.32\% & -93.98\% & 39.47\% & \textbf{81.95\%} \\
Roleplay as a patient & - & -70.56\% & -88.89\% & 90.00\% & \textbf{170.56\%} \\
Connect user to human when requested & - & 20.85\% & -21.50\% & 31.60\% & \textbf{59.28\%} \\
\midrule
Average relative improvement & - & -39.07\% & -64.51\% & 40.51\% & \textbf{86.20\%} \\
\bottomrule
\end{tabular}
\end{table}

\section{Limitations and Future Work}

Maintaining the accuracy of evaluations is critical for SAIL. Our research indicates that self-alignment checks often favor the model's own generated responses, potentially leading to the misidentification of complex cases. This bias can skew the distribution of training data, resulting in suboptimal outcomes. Additionally, although our approach excels at detecting representation gaps, it primarily focuses on augmenting responses with diversity to individual questions. This strategy is frequently insufficient for disciplines like mathematics and coding, which also require a wide range of question types. We advocate for future research to emphasize automatic data augmentation based on 'meta-logic' to improve the resilience and versatility of our techniques.

Ultimately, we aspire to develop an ideal version of the Integrated Growth and Reasoning (IGA) system as a fully autonomous self-enhancement framework for Vision-Language Models (VLMs). This system would integrate reasoning enhancements from Integrated Growth Processes (IGP) into the model's learning seamlessly. Moreover, it would use visual grounding through feedback from virtual environments to tackle the challenges in evaluation and augmentation, particularly in light of the extensive online interaction data (see \cite{putta2024agentq}). By incorporating these innovations, we aim to construct a more advanced and self-optimizing IGA system capable of adapting and evolving without human oversight.

\bibliographystyle{plainnat} 

\end{document}